\documentclass{article} 
\usepackage{iclr2026_conference,times}
\usepackage{enumitem}
\usepackage{booktabs}
\usepackage{multirow}
\usepackage{siunitx}
\usepackage{arydshln}
\usepackage{amsmath,amsfonts,bm,amsthm}
\usepackage{centernot}
\usepackage{amssymb}

\theoremstyle{definition}

\theoremstyle{remark}

\def\eqref#1{equation~\ref{#1}}

\def\1{\bm{1}}

\DeclareMathAlphabet{\mathsfit}{\encodingdefault}{\sfdefault}{m}{sl}
\SetMathAlphabet{\mathsfit}{bold}{\encodingdefault}{\sfdefault}{bx}{n}

\usepackage{hyperref}
\usepackage{url}
\usepackage{graphicx}
\usepackage{capt-of}
\title{Rethinking Classifier-Free Guidance in On-Policy Diffusion Distillation}

\author{Bingnan Li, Haozhe Wang, Haozhong Xiong, Fangtai Wu, Jinpeng Yu, Yang Shi,\\ \textbf{Jiaming Liu$^\dagger$, Ruihua Huang}\\
Qwen Business Unit of Alibaba, $^\dagger$ Corresponding Author \\
\texttt{jmliu1217@gmail.com}\\
\url{https://rethinking-cfg-opd.github.io}\\
}

\usepackage[most]{tcolorbox}
\usepackage{placeins}

\definecolor{opdblue}{RGB}{70,90,155}

\newtcolorbox{defbox}[1][]{
  enhanced,
  breakable,
  colback=opdblue!4!white,
  colframe=opdblue!28!white,
  boxrule=0.5pt,
  arc=2pt,
  borderline west={2.8pt}{0pt}{opdblue},
  left=8pt,
  right=8pt,
  top=10pt,
  bottom=6pt,
  before skip=9pt,
  after skip=9pt,
  title={#1},
  fonttitle=\bfseries\small,
  coltitle=white,
  attach boxed title to top left={
    xshift=7pt,
    yshift=-2mm
  },
  boxed title style={
    colback=opdblue,
    colframe=opdblue,
    boxrule=0pt,
    arc=1.5pt,
    left=5pt,
    right=5pt,
    top=2pt,
    bottom=2pt
  }
}

\iclrfinalcopy 
\begin{document}

\maketitle
\lhead{Preprint.}
\vspace{-2em}
\begin{center}
    \begin{minipage}{\textwidth}
        \centering
        \includegraphics[width=\textwidth]{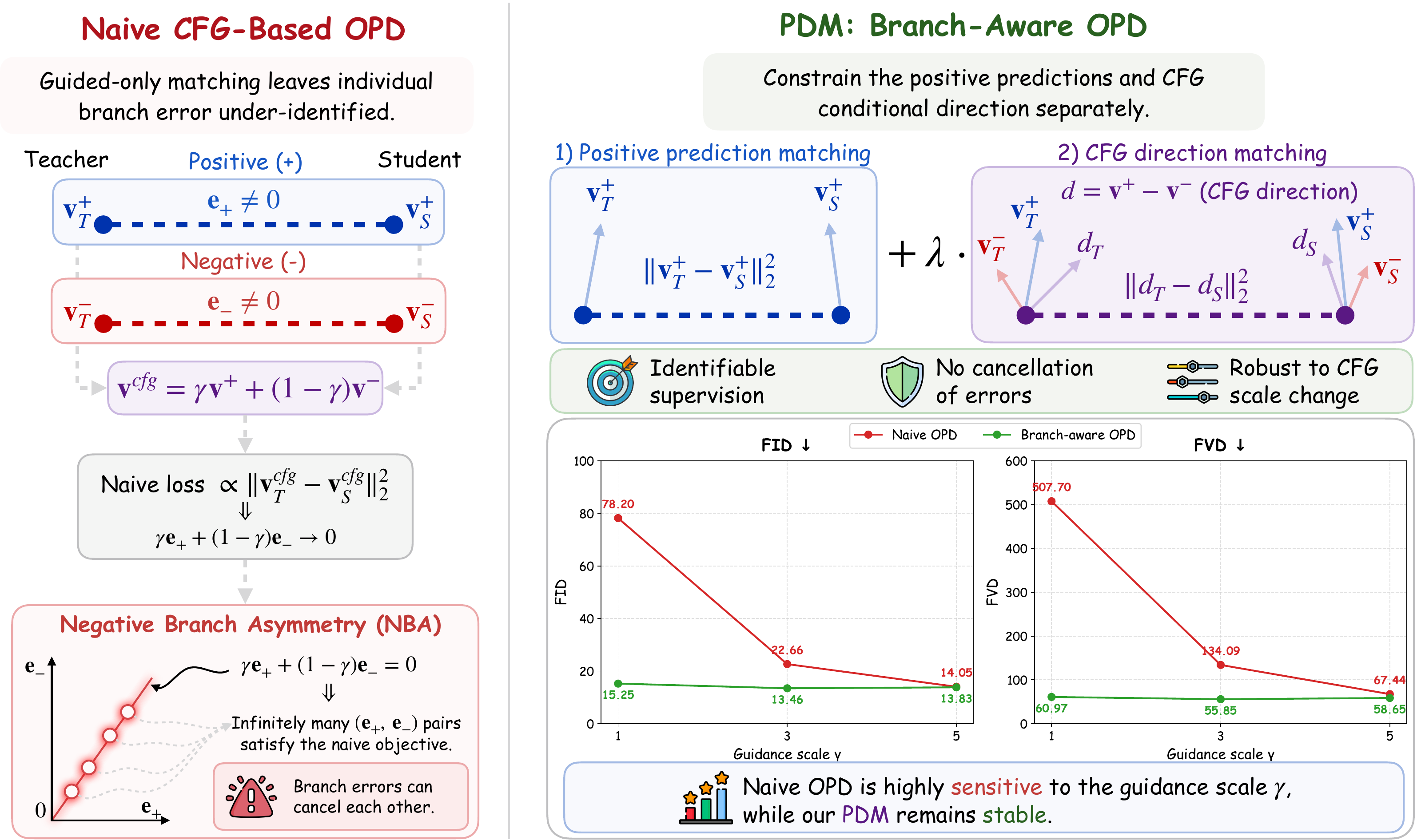}
        \captionof{figure}{
            \textbf{Naive CFG-based OPD suffers from branch ambiguity.}
Matching only the CFG-composed prediction allows positive and negative branch
errors to cancel, creating infinitely many degenerate solutions. Our
branch-aware objective, Positive--Direction Matching (PDM), resolves this ambiguity by
separately constraining the positive prediction and the CFG conditional direction,
restoring identifiable supervision and robust distillation across guidance scales.}
        \label{fig:teaser}
    \end{minipage}
\end{center}

\begin{abstract}
On-policy distillation (OPD) adapts diffusion models by querying a teacher along trajectories generated by the current student, but how it should behave under classifier-free guidance (CFG), a default component of modern diffusion systems, remains poorly understood. Existing OPD methods naturally extend velocity matching to the CFG-composed prediction, directly matching teacher and student guided velocities. We show that this objective is under-identified at the branch level: positive- and negative-branch errors can compensate in the guided prediction. Through two contrasting cases, we find that naive matching remains effective under shared negative conditioning, where both branch errors decrease jointly. When the model's native CFG schema retains privileged information in the teacher's negative branch that is unavailable to the student, however, this joint reduction breaks down and the composed objective induces antagonistic branch-error dynamics, reducing the positive-branch error while increasing the negative-branch error. We term this failure mode Negative Branch Asymmetry (NBA). To address NBA, we introduce Positive--Direction Matching (PDM), a branch-aware OPD objective that separately constrains the positive prediction and the CFG conditional direction. We apply PDM to dense-to-sparse video control, where naive guided matching is highly sensitive to inference guidance scales, while branch-aware supervision enables more robust and effective knowledge transfer.

\end{abstract}

\section{Introduction}
\label{sec:intro}

Classifier-free guidance (CFG)~\citep{cfg} drives nearly every modern diffusion
model~\citep{flux-2-2025,sd3,wan2025wan}, yet the velocity it denoises with is
never produced by the network directly. Instead, CFG evaluates the model twice:
a positive branch under the target text condition and a negative branch under
the model's null or negative text condition. It then composes the resulting
velocity predictions as
$\widetilde{\mathbf v}=\gamma\mathbf v^{+}+(1-\gamma)\mathbf v^{-}$,
where the guidance scale $\gamma$ is a deployment knob chosen at inference,
often long after training.

On-policy distillation (OPD) adapts a diffusion model by letting the student
generate a denoising trajectory with its current policy, evaluating teacher and
student at the same student-visited states, and updating the student to match
the teacher's local velocity~\citep{dopsd,flowopd,diffusionopd}. The combination
of on-policy state coverage and dense per-step supervision makes OPD an
appealing alternative to offline distillation and scalar-reward
optimization~\citep{gkd,sod,rationalrewards}.

Faced with a guided teacher, diffusion distillation can handle CFG in different
ways. Some methods absorb a CFG-defined teacher field into a single student
prediction, thereby internalizing the guidance strength used to construct the
target~\citep{luo2023latent,dmd,danceopd}. In the setting considered here,
however, both teacher and student retain the standard CFG interface of the
underlying diffusion model and produce separate positive and negative
predictions~\citep{flowopd,diffusionopd}. A natural extension of OPD is then to
match their CFG-composed velocities, $\mathcal{L}_{\mathrm{naive}}\propto
\|\widetilde{\mathbf v}^{T}-\widetilde{\mathbf v}^{S}\|_2^2.$
Because the two branches remain available, the student can still be evaluated
with different guidance scales at inference; however, the training loss
constrains only their composition at the training scale.

This composed objective is under-identified at the branch level. Let
$\mathbf e_{+}=\mathbf v_T^{+}-\mathbf v_S^{+}$ and
$\mathbf e_{-}=\mathbf v_T^{-}-\mathbf v_S^{-}$ denote the positive- and
negative-branch errors. Guided matching constrains only
$\gamma\mathbf e_{+}+(1-\gamma)\mathbf e_{-}$, allowing the two errors to
compensate without changing the composed prediction. Branch ambiguity alone does not imply failure. Under shared negative
conditioning, shared-parameter updates may reduce both branch errors jointly.
Under privileged negative conditioning, however, our experiments show that
this joint reduction can break down, exposing the compensation freedom of the
composed objective.

To examine how negative-branch conditioning affects OPD optimization, we track
the positive- and negative-branch errors throughout training in two
image-domain distillation settings. In text-rendering distillation~\citep{diffusionopd}, both errors
decrease jointly and naive matching remains effective. In
reference-conditioned distillation~\citep{dopsd}, where the teacher's negative branch
receives privileged information, naive matching instead reduces the
positive-branch error while increasing the negative-branch error, inducing
\emph{antagonistic branch-error dynamics}. We term this failure mode
\emph{Negative Branch Asymmetry} (NBA).

NBA can remain hidden at the guidance scale used for training because the
opposing branch errors compensate in the composed prediction. Once the guidance
scale changes at inference, the branches are recomposed with different weights
and the learned compensation no longer holds. Guidance-scale sensitivity is
therefore a direct observable consequence of NBA: a student can closely match
the teacher at $\gamma_{\mathrm{train}}$ yet drift when evaluated at another
guidance scale. Figure~\ref{fig:teaser} illustrates this branch ambiguity and
its effect under guidance-scale changes.

The remedy is to supervise before composition. Our primary objective,
\emph{Positive--Direction Matching} (PDM), separately constrains the positive
prediction and the CFG conditional direction
$\mathbf v^{+}-\mathbf v^{-}$. Zero PDM loss forces both branch errors to zero,
preventing the composed objective from improving one branch at the expense of
the other. As a foil, we also study \emph{Independent Branch Matching} (IBM),
which directly matches the positive and negative branches. Both objectives
remove the compensation freedom of guided-only matching, while weighting
branch-level supervision differently.

Having isolated when NBA occurs, we apply branch-aware OPD to dense-to-sparse
video control on Wan-VACE~\citep{vace}. The teacher observes dense per-frame
control, whereas the student receives only sparse keyframes. In this practical
setting, naive guided matching degrades sharply when the inference guidance
scale departs from its training value, while PDM and IBM remain stable. PDM also
improves control fidelity across pose, depth, and scribble modalities,
demonstrating that branch-aware supervision enables more robust and effective
knowledge transfer beyond the image-domain case study.

Our contributions are summarized as follows:
\begin{itemize}
[leftmargin=*,topsep=0pt,itemsep=0pt,parsep=0pt,partopsep=0pt]
    \item We show that CFG-composed OPD is under-identified at the branch level and identify when this ambiguity becomes harmful. It remains benign under shared negative conditioning, but produces NBA when privileged negative conditioning breaks joint branch-error reduction and induces antagonistic branch-error dynamics.

    \item We introduce \emph{Positive--Direction Matching} (PDM), a branch-aware
    OPD objective that supervises the positive prediction and the CFG
    conditional direction separately, preventing cross-branch error
    compensation and improving robustness across inference guidance scales. We
    include Independent Branch Matching (IBM) as a branch-aware foil.

    \item We apply branch-aware OPD to dense-to-sparse video control and provide
    quantitative evidence across pose, depth, and scribble modalities. Naive
    matching exhibits severe guidance-scale sensitivity, while branch-aware
    supervision transfers dense control knowledge more reliably.
\end{itemize}

\section{Preliminaries}
\label{sec:preliminaries}

\subsection{On-Policy Distillation for Diffusion Models}
\label{sec:prelim_opd}

Diffusion on-policy distillation (OPD) trains a student model using states
sampled from its own evolving generation process
\citep{dopsd,flowopd,diffusionopd}. Let
\begin{equation}
    \mathbf{x}_t^S \sim p_{\theta}
\end{equation}
denote a noisy state visited along a trajectory generated by the current
student. At this state, the teacher and student define local denoising
transitions conditioned on $\mathbf{c}_T$ and $\mathbf{c}_S$, respectively.
The teacher may have access to privileged information unavailable or only
partially available to the student, and therefore $\mathbf{c}_T$ and
$\mathbf{c}_S$ need not be identical.

Existing diffusion OPD formulations
\citep{dopsd,flowopd,diffusionopd} minimize the discrepancy between teacher and
student transitions at student-visited states. Although they differ in their
derivations and optimization procedures, their transition-level objectives
reduce, up to a timestep-dependent weighting factor, to matching the
corresponding model predictions:
\begin{equation}
    \mathcal{L}_{\mathrm{OPD}}
    =
    \mathbb{E}_{\mathbf{x}_t^S \sim p_{\theta},\,t}
    \left[
        w(t)
        \left\|
            \mathbf{v}_{T}
            \left(\mathbf{x}_t^S,t,\mathbf{c}_T\right)
            -
            \mathbf{v}_{S}
            \left(\mathbf{x}_t^S,t,\mathbf{c}_S\right)
        \right\|_2^2
    \right],
    \label{eq:opd_loss}
\end{equation}
where $\mathbf{v}_{T}$ and $\mathbf{v}_{S}$ denote the teacher and student
velocity predictions, and $w(t)$ absorbs the parameterization- and
solver-dependent weighting. We use velocity prediction throughout the paper,
while the same formulation applies to equivalent noise, score, or flow
parameterizations.

\subsection{Diffusion OPD with Classifier-Free Guidance}
\label{sec:prelim_cfg}

We next consider diffusion models that use classifier-free guidance (CFG) with
guidance scale $\gamma>1$. The velocity used for denoising is
\begin{equation}
    \widetilde{\mathbf{v}}
    =
    \gamma\mathbf{v}^{+}
    +
    (1-\gamma)\mathbf{v}^{-},
    \label{eq:cfg_prediction}
\end{equation}
where $\mathbf{v}^{+}$ and $\mathbf{v}^{-}$ denote the positive and negative branch predictions, respectively.
We preserve the CFG schema
prescribed by each model's original training and inference pipeline. In many models, the negative branch
changes the text condition while retaining auxiliary inputs such as reference
images or other control signals.

Define the positive- and negative-branch discrepancies as
\begin{equation}
    \mathbf{e}_{+}
    =
    \mathbf{v}_{T}^{+}
    -
    \mathbf{v}_{S}^{+},
    \qquad
    \mathbf{e}_{-}
    =
    \mathbf{v}_{T}^{-}
    -
    \mathbf{v}_{S}^{-}.
    \label{eq:branch_errors}
\end{equation}
Substituting these discrepancies into the CFG-composed matching objective gives
\begin{equation}
    \mathcal{L}_{\mathrm{naive}}
    =
    \mathbb{E}_{\mathbf{x}_t^S \sim p_{\theta},\,t}
    \left[
        w(t)
        \left\|
            \gamma\mathbf{e}_{+}
            +(1-\gamma)\mathbf{e}_{-}
        \right\|_2^2
    \right].
    \label{eq:cfg_opd_loss}
\end{equation}

Existing branch-retaining OPD settings~\citep{flowopd,diffusionopd} are
primarily studied in text-to-image distillation, where teacher and student
share the same null-text conditioning on the negative branch. Such symmetry is
not guaranteed in OPD with privileged information, where the teacher may
receive reference or control information unavailable or only partially
available to the student. Because the model's native CFG schema may retain this
information in the negative branch, privileged negative conditioning arises
naturally in practical knowledge-transfer settings. We analyze its consequence
next.

\begin{figure}[t]
    \centering
    \includegraphics[width=1\linewidth]{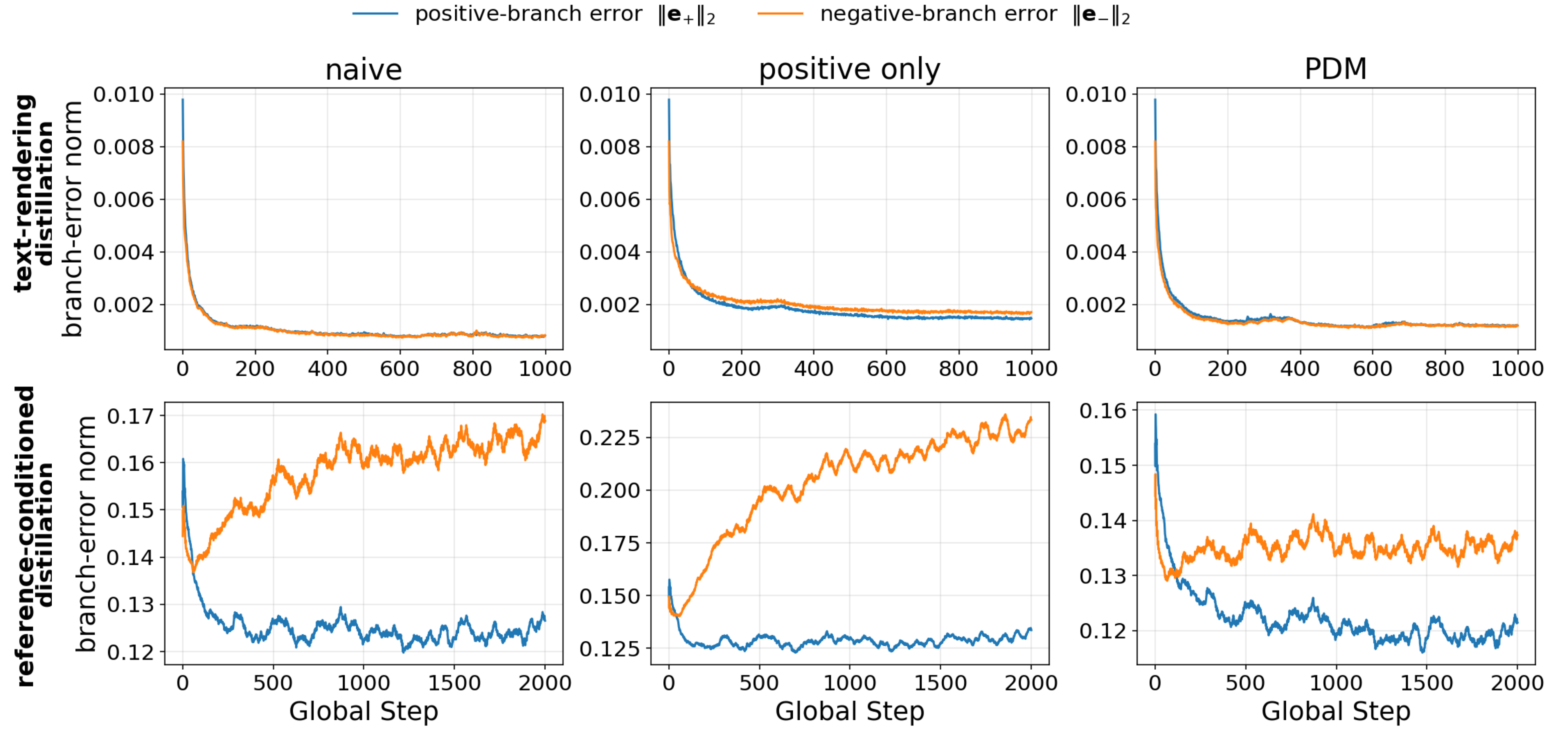}
\caption{\textbf{Branch-error dynamics under shared and privileged negative conditioning.} We track the positive- and negative-branch $\ell_2$ error norms during text-rendering (top) and reference-conditioned (bottom) distillation. Under shared negative conditioning, both errors decrease jointly for all objectives. With privileged reference conditioning, naive and positive-only matching reduce the positive error while increasing the negative error, exhibiting antagonistic branch-error dynamics. PDM suppresses this negative-branch degradation by explicitly constraining both the positive prediction and CFG conditional direction.}
    \label{fig:branch_dynamics}
\end{figure}

\section{Negative Branch Asymmetry and Branch-Aware OPD}
\label{sec:method}

\subsection{Negative Branch Asymmetry}
\label{sec:nba}

Under privileged negative conditioning, the teacher's negative target cannot
be directly reproduced from the student's negative input. Nevertheless,
Eq.~\ref{eq:cfg_opd_loss} constrains only the CFG-composed discrepancy. At any
supervised state, matching the guided predictions requires only
\begin{equation}
    \gamma\mathbf{e}_{+}
    +(1-\gamma)\mathbf{e}_{-}
    =
    \mathbf{0},
\end{equation}
or equivalently,
\begin{equation}
    \mathbf{e}_{+}
    =
    \frac{\gamma-1}{\gamma}\mathbf{e}_{-}.
    \label{eq:cancellation}
\end{equation}
This condition admits infinitely many non-zero branch-error pairs, allowing the
two errors to compensate without changing the guided prediction.

Equation~\ref{eq:cancellation}
establishes the branch ambiguity but does not determine its optimization dynamics.
We therefore investigate when this ambiguity becomes harmful empirically in
Section~\ref{sec:nba_diagnosis}. Across two controlled image-domain cases,
we observe joint branch-error reduction under shared negative conditioning,
but antagonistic branch-error dynamics under privileged negative conditioning.

\begin{defbox}{Definition: Negative Branch Asymmetry (NBA).} Branch ambiguity alone does not constitute a failure.
We define NBA as the failure
mode in which, under privileged negative conditioning, naive CFG-composed
matching induces antagonistic branch-error dynamics, reducing
$\|\mathbf e_{+}\|_2$ while increasing $\|\mathbf e_{-}\|_2$.
The opposing errors can remain hidden at the training guidance scale and become
exposed when the branches are recomposed at another scale.
\end{defbox}

An absolute performance drop under a guidance-scale change is not by itself
evidence of NBA, since the teacher may intrinsically rely on CFG. The relevant
signature is excess degradation introduced by naive distillation: the student
departs from the teacher's guidance-scale behavior or degrades more strongly
than its branch-aware counterpart. Section~\ref{sec:nba_diagnosis} tests these
predictions empirically.

\begin{figure}[t]
    \centering
    \includegraphics[width=\linewidth]{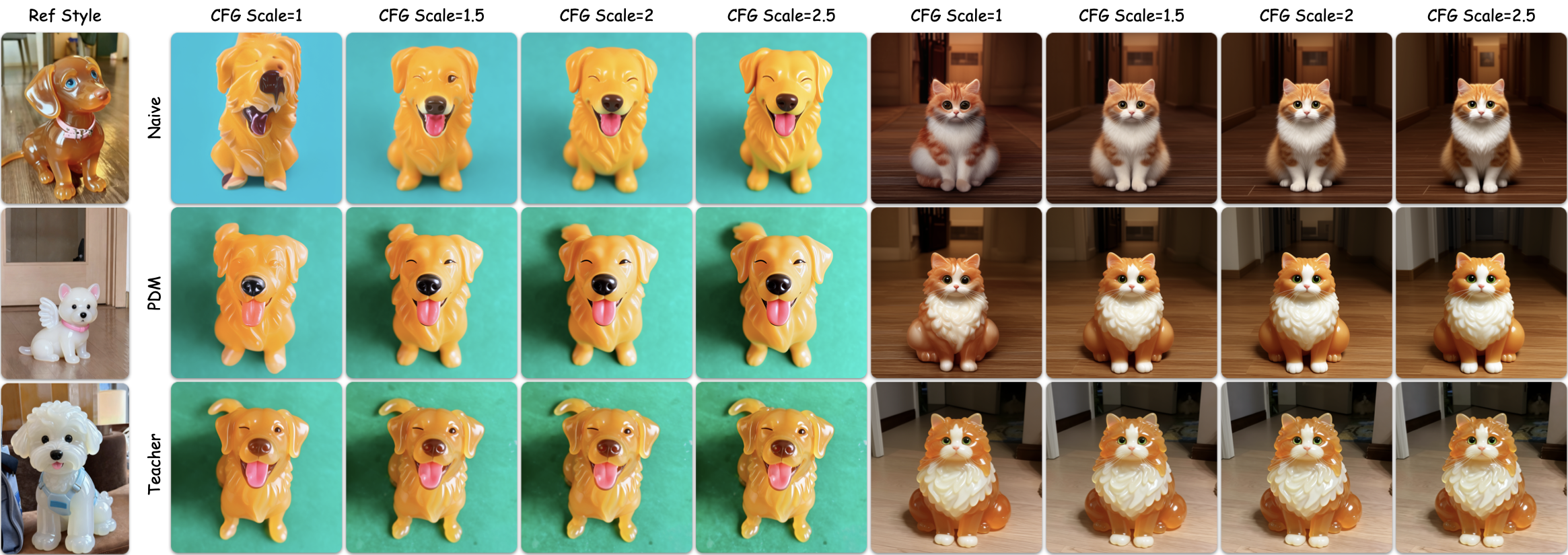}
\caption{\textbf{Distillation-induced CFG sensitivity under privileged reference conditioning.} The left column shows the reference-style exemplars. For two held-out prompts, we compare the reference-conditioned teacher with text-only students trained using naive matching or PDM at $\gamma_{\mathrm{train}}=2$. All outputs use matched prompts and initial noise. The teacher preserves the reference-supplied style across inference guidance scales, and PDM closely follows this behavior. Naive matching instead exhibits stronger visual distortion and style drift, most notably at $\gamma=1$. Additional examples are provided in Appendix~\ref{app:more_qual_res}.}
    \label{fig:reference_cfg}
\end{figure}

\subsection{Branch-Aware OPD}
\label{sec:branch_aware_objectives}

To prevent cross-branch error compensation, we supervise the predictions before
CFG composition. Let $\ell$ denote a pointwise distillation loss evaluated at a
student-visited state $\mathbf{x}_t^S$ and timestep $t$. Its corresponding OPD
objective is
\begin{equation}
    \mathcal{L}_{\mathrm{OPD}}[\ell]
    =
    \mathbb{E}_{\mathbf{x}_t^S\sim p_\theta,\,t}
    \left[
        w(t)\,\ell(\mathbf{x}_t^S,t)
    \right].
    \label{eq:generic_opd_objective}
\end{equation}

\paragraph{Positive--Direction Matching.}
For a model $M\in\{T,S\}$, define its CFG conditional direction as
\begin{equation}
    \mathbf{d}_M
    =
    \mathbf{v}_M^{+}-\mathbf{v}_M^{-}.
    \label{eq:conditional_direction}
\end{equation}
Our primary objective, Positive--Direction Matching (PDM), separately matches
the positive prediction and the conditional direction:
\begin{equation}
    \ell_{\mathrm{PDM}}
    =
    \left\|\mathbf{e}_{+}\right\|_2^2
    +
    \lambda
    \left\|\mathbf{d}_T-\mathbf{d}_S\right\|_2^2
    =
    \left\|\mathbf{e}_{+}\right\|_2^2
    +
    \lambda
    \left\|\mathbf{e}_{+}-\mathbf{e}_{-}\right\|_2^2,
\label{eq:pdm}
\end{equation}
where $\lambda>0$ controls the direction-matching strength. The first term
anchors the positive prediction, while the second preserves the CFG
conditional direction. Zero PDM loss requires
$\mathbf{e}_{+}=\mathbf{e}_{-}=\mathbf{0}$, eliminating the compensation
freedom of guided-only matching.

\paragraph{Baseline: Independent Branch Matching.}
To separate the benefit of branch-level supervision from the particular design
of PDM, we include Independent Branch Matching (IBM) as a baseline:
\begin{equation}
    \ell_{\mathrm{IBM}}
    =
    \gamma^2\left\|\mathbf{e}_{+}\right\|_2^2
    +
    (\gamma-1)^2\left\|\mathbf{e}_{-}\right\|_2^2.
    \label{eq:ibm}
\end{equation}
IBM retains the per-branch weights of the expanded naive objective but removes
its cross-branch interaction term. Like PDM, it has a unique branch-level
zero-loss solution for $\gamma>1$, but directly matches the two branches rather
than preserving the positive prediction and conditional direction. We compare the empirical behavior of these two branch-aware formulations in Table~\ref{tab:main}.

\begin{table}[t]
    \centering
    \caption{\textbf{Intrinsic CFG sensitivity in text-rendering distillation.}
    OCR reward (\%, $\uparrow$) on 1,018 held-out prompts. All models exhibit a
    similar response to the inference guidance scale, showing that the drop at
    $\gamma=1$ is inherited from the teacher rather than introduced by naive
    distillation.}
    \label{tab:ocr_cfg}
    \small
    \setlength{\tabcolsep}{8pt}
    \begin{tabular}{lccccc}
        \toprule
        Method
        & $\gamma{=}1$
        & $\gamma{=}2$
        & $\gamma{=}3$
        & $\gamma{=}4$
        & $\gamma{=}4.5$ \\
        \midrule
        Teacher     & 75.24 & 92.91 & 94.16 & 94.43 & 94.09 \\
        OPD (Naive) & 73.87 & 92.72 & 94.52 & 94.43 & 94.11 \\
        OPD (PDM)   & 74.48 & 93.73 & 93.99 & 94.02 & 94.38 \\
        \bottomrule
    \end{tabular}
\end{table}

\subsection{Efficient Supervision for Dense-to-Sparse Video Control}
\label{sec:partial_trajectory}

Full-trajectory OPD is particularly expensive for video because it requires a
teacher evaluation at every denoising state. For our dense-to-sparse video
application, we independently find that supervising only an early portion of the student trajectory substantially
reduces this cost while maintaining performance.

Let $\{\mathbf{x}_{t_k}^{S}\}_{k=0}^{N-1}$ denote the states visited by the
student in denoising order. We supervise only the first $K$ states:
\begin{equation}
    \mathcal{L}_{\mathrm{OPD}}^{(K)}
    =
    \mathbb{E}_{\{\mathbf{x}_{t_k}^{S}\}_{k=0}^{N-1}\sim p_\theta}
    \left[
        \frac{1}{K}
        \sum_{k=0}^{K-1}
        w(t_k)\,
        \ell(\mathbf{x}_{t_k}^{S},t_k)
    \right].
    \label{eq:partial_trajectory_opd}
\end{equation}
The student still completes the full denoising rollout, but teacher supervision
is computed only for these $K$ states. Here $K=N$ recovers full-trajectory
supervision, while smaller $Kx$ reduces training cost. This design is used only
for the dense-to-sparse video experiments; the image-domain case studies follow
their respective standard supervision protocols. We study the effect of $K$ in
Section~\ref{sec:opd_horizon}.

\begin{table}[t]
\centering
\small
\setlength{\tabcolsep}{4pt}
\setlength{\dashlinedash}{1.5pt}
\setlength{\dashlinegap}{2pt}
\caption{\textbf{Main results on dense-to-sparse video control.}
Models are jointly trained on pose, depth, and scribble controls and evaluated
separately on each modality at the training guidance scale $\gamma=5$.
\textbf{Bold} and \underline{underlined} values denote the best and second-best
results, respectively, excluding the teacher.} 

\label{tab:main}

\begin{tabular*}{\textwidth}{@{\extracolsep{\fill}}lccccccccc}
\toprule
Pose-Control
& \multicolumn{2}{c}{MPJPE $\downarrow$}
& \multicolumn{2}{c}{PCK@0.2 $\uparrow$}
& \multicolumn{2}{c}{PCK@0.1 $\uparrow$}
& \multicolumn{2}{c}{Quality $\downarrow$} \\
\cmidrule(lr){2-3}\cmidrule(lr){4-5}\cmidrule(lr){6-7}\cmidrule(lr){8-9}

Method
& All & Key
& All & Key
& All & Key
& FID & FVD \\
\midrule
Teacher & 3.03 & 2.82 & 96.02 & 96.46 & 91.15 & 92.23 & 14.13 & 56.01 \\
Student & 5.92 & 4.10 & 86.78 & 91.89 & 69.97 & 82.53 & 15.81 & 81.41 \\

\midrule
SFT & 5.02 & 3.31 & 90.36 & 94.77 & 76.72 & 88.55 & 14.11 & \underline{63.90} \\
Off-policy & 5.14 & 3.44 & 89.78 & 94.09 & 75.46 & 87.48 & 14.49 & 68.50 \\

\midrule
OPD (Naive) & 4.43 & 3.14 & 91.84 & 95.25 & 80.28 & 89.35 & \underline{13.49} & 65.12 \\
OPD (IBM) & \underline{4.20} & \underline{2.88} & \underline{92.93} & \underline{96.11} & \underline{82.27} & \underline{90.95} & \textbf{13.26} & 65.72 \\
OPD (PDM) & \textbf{4.13} & \textbf{2.83} & \textbf{92.98} & \textbf{96.12} & \textbf{82.48} & \textbf{91.03} & 13.52 & \textbf{62.77} \\
\midrule
\midrule
Depth-Control
& \multicolumn{2}{c}{CORR $\uparrow$}
& \multicolumn{2}{c}{RMSE $\downarrow$}
& \multicolumn{2}{c}{SI-RMSE $\downarrow$}
& \multicolumn{2}{c}{Quality $\downarrow$} \\
\cmidrule(lr){2-3}\cmidrule(lr){4-5}\cmidrule(lr){6-7}\cmidrule(lr){8-9}

Method
& All & Key
& All & Key
& All & Key
& FID & FVD \\
\midrule

Teacher & 96.98 & 97.58 & 7.50 & 6.36 & 6.19 & 5.23 & 11.02 & 36.89 \\
Student & 83.21 & 91.27 & 16.87 & 11.13 & 14.36 & 9.37 & 15.03 & 71.33 \\

\midrule
SFT & 88.39 & 96.17 & 13.83 & 7.80 & 11.96 & 6.52 & 12.87 & \textbf{58.13} \\
Off-policy & 88.25 & 96.01 & 13.98 & 8.00 & 12.10 & 6.64 & 12.66 & \underline{59.35} \\

\midrule
OPD (Naive) & 90.58 & 96.56 & 12.39 & 7.47 & 10.69 & 6.16 & 12.08 & 72.20\\
OPD (IBM) & \underline{91.14} & \underline{96.91} & \underline{12.03} & \underline{7.11} & \underline{10.37} & \underline{5.85} & \underline{11.76} & 60.38 \\
OPD (PDM) & \textbf{91.24} & \textbf{97.13} & \textbf{11.95} & \textbf{6.91} & \textbf{10.31} & \textbf{5.68} & \textbf{11.64} & 60.60 \\

\midrule
\midrule
Scribble-Control
& \multicolumn{2}{c}{Recall $\uparrow$}
& \multicolumn{2}{c}{Precision $\uparrow$}
& \multicolumn{2}{c}{F1 $\uparrow$}
& \multicolumn{2}{c}{Quality $\downarrow$} \\
\cmidrule(lr){2-3}\cmidrule(lr){4-5}\cmidrule(lr){6-7}\cmidrule(lr){8-9}

Method
& All & Key
& All & Key
& All & Key
& FID & FVD \\
\midrule

Teacher & 93.26 & 94.40 & 94.26 & 95.27 & 93.73 & 94.81 & 9.56 & 24.22 \\
Student & 64.20 & 82.93 & 65.50 & 83.78 & 64.70 & 83.24 & 12.29 & 69.13 \\

\midrule
SFT & 67.23 & 89.03 & 69.52 & 90.66 & 68.27 & 89.79 & 10.69 & \underline{50.56} \\
Off-policy & 70.98 & 91.55 & 72.15 & 91.73 & 71.49 & 91.60 & 11.60 & 52.90 \\

\midrule
OPD (Naive) & 74.78 & 90.77 & 75.38 & 91.34 & 75.00 & 90.99 & 10.53 & 59.95\\
OPD (IBM) & \underline{75.92} & \underline{92.23} & \textbf{76.47} & \underline{92.70} & \underline{76.13} & \underline{92.43} & \textbf{10.12} & 52.67 \\
OPD (PDM) & \textbf{76.42} & \textbf{93.03} & \underline{76.13} & \textbf{92.86} & \textbf{76.22} & \textbf{93.03} & \underline{10.34} & \textbf{50.53} \\

\bottomrule
\end{tabular*}
\end{table}

\section{Experiments}
\label{sec:experiments}

We first investigate when the branch ambiguity of CFG-composed matching
develops into NBA. We then evaluate branch-aware OPD in dense-to-sparse video
control and ablate its main design choices.

\subsection{When Does NBA Emerge?}
\label{sec:nba_diagnosis}

\subsubsection{Experimental Setup}

We contrast two image-domain distillation settings with different negative
conditioning. In text-rendering distillation, we follow DiffusionOPD~\citep{diffusionopd} and use an SD3.5-Medium-based~\citep{sd3} OCR teacher
specialized for visual text rendering. Teacher and student receive the same
prompt on the positive branch and the same null-text condition on the negative
branch:
\begin{equation}
    \mathbf{c}_T^{+}=\mathbf{c}_S^{+}=\mathbf{y},
    \qquad
    \mathbf{c}_T^{-}=\mathbf{c}_S^{-}=\varnothing.
\end{equation}
Models are trained at $\gamma_{\mathrm{train}}=4.5$ and evaluated on 1,018
held-out prompts.

In reference-conditioned distillation~\citep{dopsd}, we conduct experiments
with FLUX-2-klein-base-4B~\citep{flux-2-2025}, with the teacher observing a
reference image $\mathbf r$ unavailable to the text-only student:
\begin{equation}
\begin{aligned}
    \mathbf{c}_T^{+}&=(\mathbf{y},\mathbf{r}),
    &\mathbf{c}_S^{+}&=\mathbf{y},\\
    \mathbf{c}_T^{-}&=(\varnothing,\mathbf{r}),
    &\mathbf{c}_S^{-}&=\varnothing.
\end{aligned}
\end{equation}
The teacher therefore receives privileged information on its negative branch.
Models are trained at $\gamma_{\mathrm{train}}=2$ and evaluated on held-out
prompts across inference guidance scales.

To diagnose branch coupling, we train naive matching, PDM, and a positive-only
ablation ($\ell_{+}=\|\mathbf e_{+}\|_2^2$),
and track both branch errors throughout training. The positive-only ablation
isolates how updates toward the positive target affect the negative-branch
error. It is used only for this optimization diagnostic; generation comparisons
use naive matching and PDM under the same training protocol. Full implementation
details are provided in
Appendix~\ref{app:setup}.

\subsubsection{Branch-Error Dynamics}

Figure~\ref{fig:branch_dynamics} reveals two distinct optimization regimes. In
text-rendering distillation, positive-only training reduces both branch errors,
showing that positive-branch updates also improve the negative branch. Naive
matching and PDM exhibit the same joint error reduction, indicating a benign
regime.

Reference-conditioned distillation exhibits the opposite behavior.
Positive-only training reduces the positive-branch error but substantially
increases the negative-branch error, showing that positive-branch updates no
longer benefit the negative branch under privileged conditioning. Naive
matching follows the same antagonistic pattern, decreasing the positive error
while increasing the negative error. PDM instead reduces the positive error
while preventing sustained negative-error growth. The antagonistic trajectory
under naive matching is the optimization signature of NBA, while PDM restores
joint branch-level supervision.

\begin{table}[t]
\centering
\small
\setlength{\tabcolsep}{4pt}
\caption{\textbf{Guidance-scale generalization on pose control.}
Models are trained at $\gamma_{\mathrm{train}}=5$ and evaluated at different
inference guidance scales.}
\label{tab:cfg_ablation}

\begin{tabular*}{\textwidth}{@{\extracolsep{\fill}}lccccccccc}
\toprule

& \multicolumn{2}{c}{MPJPE $\downarrow$}
& \multicolumn{2}{c}{PCK@0.2 $\uparrow$}
& \multicolumn{2}{c}{PCK@0.1 $\uparrow$}
& \multicolumn{2}{c}{Quality $\downarrow$} \\
\cmidrule(lr){2-3}\cmidrule(lr){4-5}\cmidrule(lr){6-7}\cmidrule(lr){8-9}

Method
& All & Key
& All & Key
& All & Key
& FID & FVD \\

\midrule

Naive ($\gamma=5$) & 4.62 & 3.22 & 91.35 & 95.01 & 79.39 & 89.54 & 14.05 & 67.44 \\
Naive ($\gamma=3$) & 4.94 & 3.31 & 90.23 & 94.53 & 77.92 & 88.22 & 22.66 & 134.09 \\
Naive ($\gamma=1$) & 8.98 & 6.23 & 80.08 & 87.42 & 69.14 & 81.57 & 78.20 & 507.70 \\

\midrule
IBM ($\gamma=5$) & 4.28 & 2.91 & 92.67 & 95.95 & 81.69 & 91.03 & 12.99 & 53.83 \\
IBM ($\gamma=3$) & 4.42 & 3.04 & 92.09 & 95.53 & 81.41 & 90.81 & 13.69 & 56.85 \\
IBM ($\gamma=1$) & 4.91 & 3.43 & 90.92 & 94.69 & 80.66 & 89.98 & 18.00 & 71.17 \\

\midrule
PDM ($\gamma=5$) & 4.27 & 2.75 & 92.71 & 96.38 & 82.12 & 91.66 & 13.83 & 58.65 \\
PDM ($\gamma=3$) & 4.17 & 2.73 & 92.81 & 96.61 & 82.27 & 91.81 & 13.46 & 55.85 \\
PDM ($\gamma=1$) & 4.48 & 2.90 & 92.14 & 95.64 & 81.72 & 91.15 & 15.25 & 60.97 \\

\bottomrule
\end{tabular*}
\end{table}

\subsubsection{Intrinsic and Distillation-Induced CFG Sensitivity}

The branch-error trajectories above distinguish benign joint reduction from
NBA during optimization. We next examine how these regimes manifest under
inference-time guidance shifts. An absolute performance drop away from the
training guidance scale is not sufficient evidence of NBA, since the teacher
itself may intrinsically rely on CFG. Table~\ref{tab:ocr_cfg} shows that, in text-rendering distillation, both
students closely track the teacher throughout the guidance sweep. At
$\gamma=1$, the teacher, naive student, and PDM student obtain OCR rewards of
$75.24$, $73.87$, and $74.48$, respectively, and all improve similarly as
guidance increases. Thus, neither objective introduces substantial sensitivity
beyond the teacher's intrinsic CFG dependence.

Reference-conditioned distillation exhibits a different pattern. Figure~\ref{fig:reference_cfg} shows that the teacher preserves the reference-supplied style across guidance scales and that PDM closely tracks this behavior. Naive matching instead departs from the teacher, producing visual distortion and style drift that become most pronounced at $\gamma=1$, where CFG reduces to the positive prediction. This excess degradation relative to both the teacher and PDM isolates the sensitivity introduced by naive distillation. Together with the branch-error trajectories in Figure~\ref{fig:branch_dynamics}, these results show that NBA induces guidance-scale sensitivity beyond the teacher's intrinsic reliance on CFG.

\begin{defbox}{Finding: When Does Branch Ambiguity Become Harmful?}
Under shared negative conditioning, both branch errors decrease jointly and the
student follows the teacher's intrinsic CFG response. Under privileged negative
conditioning, naive matching induces antagonistic branch-error dynamics and
excess guidance-scale sensitivity, while PDM suppresses both. Thus, excess
degradation relative to the teacher or a branch-aware student---rather than an
absolute CFG-free performance drop---is the diagnostic signature of NBA.
\end{defbox}

\subsection{Application to Dense-to-Sparse Video Control}
\label{sec:video_application}

\subsubsection{Experimental Setup}

We apply branch-aware OPD to dense-to-sparse video control using
Wan-VACE~\citep{vace}. The teacher receives a dense control sequence
$\mathbf{p}_{\mathrm{dense}}$, while the student observes only sparse keyframes
$\mathbf{p}_{\mathrm{sparse}}$:
\begin{equation}
\begin{aligned}
    \mathbf{c}_T^{+}
    &=(\mathbf{y},\mathbf{p}_{\mathrm{dense}}),
    &
    \mathbf{c}_S^{+}
    &=(\mathbf{y},\mathbf{p}_{\mathrm{sparse}}),\\
    \mathbf{c}_T^{-}
    &=(\varnothing,\mathbf{p}_{\mathrm{dense}}),
    &
    \mathbf{c}_S^{-}
    &=(\varnothing,\mathbf{p}_{\mathrm{sparse}}).
\end{aligned}
\label{eq:d2s_conditions}
\end{equation}

We construct a benchmark from
OpenHumanVid~\citep{li2025openhumanvid} containing 600 test clips and evaluate
pose, depth, and scribble control under the dense-to-four-keyframe setting. Baselines include the unadapted sparse-control student, SFT on data-derived
diffusion states, and off-policy teacher matching; all methods use the same
student architecture and training budget.

Unless otherwise specified, PDM uses $\lambda=1$ and supervises the first
$K=8$ student-visited states. Main results use joint training over all three
control modalities. Guidance-scale and hyperparameter studies use pose-only
models to reduce repeated training cost; results from the two protocols are
therefore not directly comparable. Complete data, metric, and implementation
details are provided in Appendix~\ref{app:setup}.

\subsubsection{Main Results}

Table~\ref{tab:main} reports results at the training guidance scale
$\gamma=5$. PDM provides the most consistent control-fidelity improvements
across pose, depth, and scribble, while IBM also outperforms naive matching on
most control metrics. The performance of IBM supports the benefit of removing
cross-branch compensation, while the stronger overall results of PDM support
its positive--direction parameterization.

These matched-scale results establish the practical value of branch-aware OPD
independently of guidance-scale shifts. At
$\gamma=\gamma_{\mathrm{train}}$, PDM improves control fidelity over naive OPD,
SFT, and off-policy distillation across all three modalities. Thus,
branch-aware on-policy supervision improves dense-to-sparse knowledge transfer
itself, rather than merely correcting off-scale degradation.

\subsubsection{Guidance-Scale Generalization}

Table~\ref{tab:cfg_ablation} evaluates pose-only models trained at
$\gamma_{\mathrm{train}}=5$ and tested at different guidance scales. Naive
matching degrades sharply as the inference scale moves away from training,
reaching its largest drop at $\gamma=1$. In contrast, PDM and IBM remain stable
across the sweep, showing that branch-aware supervision prevents the excess
guidance sensitivity induced by composed matching. PDM provides the most
consistent control fidelity across the evaluated scales.

\begin{figure}[t]
    \centering
    \includegraphics[width=\linewidth]{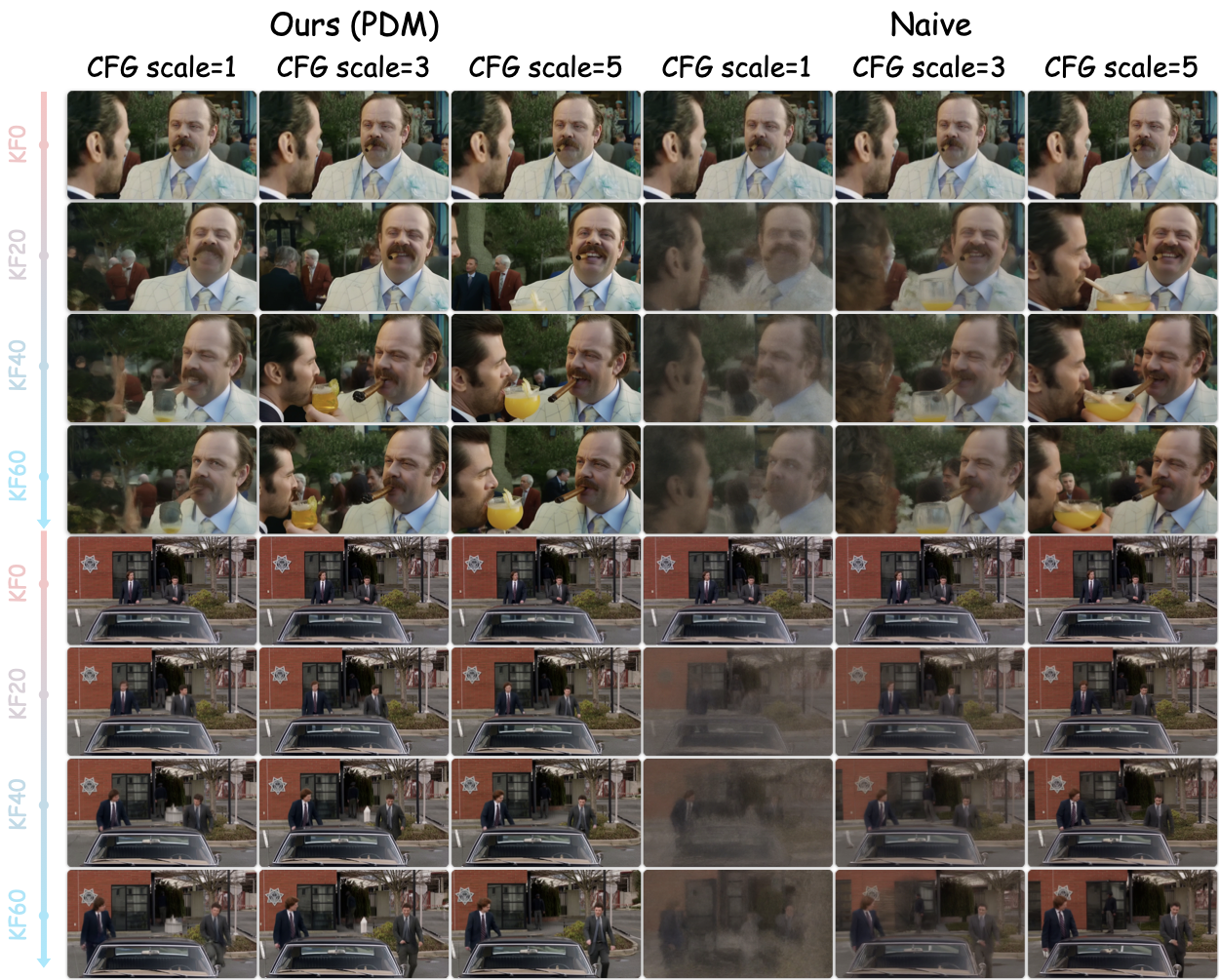}
\caption{\textbf{Guidance-scale robustness in dense-to-sparse video control.}
Models are trained at $\gamma=5$. PDM remains stable across inference scales,
whereas naive OPD degrades off-scale.}
    \label{fig:NBA_d2s}
\end{figure}

Figure~\ref{fig:NBA_d2s}  qualitatively illustrates the excess guidance-scale
sensitivity of naive OPD expected under NBA. Additional qualitative comparisons on pose, depth, and scribble-controlled generation are provided in Appendix~\ref{app:more_qual_res}.

\subsection{Ablation Studies}
\label{sec:ablation}

\subsubsection{Effect of the Direction-Matching Weight}

\begin{table}[t]
\centering
\small
\setlength{\tabcolsep}{4pt}

\caption{\textbf{Effect of the conditional-direction weight \(\lambda\) in the pose-only ablation setting.} \textbf{Bold} and \underline{underlined} numbers denote the best and
second-best results, respectively.}
\label{tab:lambda}

\begin{tabular*}{\textwidth}{@{\extracolsep{\fill}}l
cc
cc
cc
cc}
\toprule

& \multicolumn{2}{c}{MPJPE $\downarrow$}
& \multicolumn{2}{c}{PCK@0.2 $\uparrow$}
& \multicolumn{2}{c}{PCK@0.1 $\uparrow$}
& \multicolumn{2}{c}{Quality $\downarrow$} \\
\cmidrule(lr){2-3}\cmidrule(lr){4-5}\cmidrule(lr){6-7}\cmidrule(lr){8-9}

Method
& All & Key
& All & Key
& All & Key
& FID & FVD \\
\midrule
PDM ($\lambda=0$) & \textbf{4.24} & \underline{2.81} & \textbf{92.71} & \underline{96.24} & \underline{81.80} & \underline{91.05} & 13.97 & 59.72 \\
PDM ($\lambda=1$) & \underline{4.27} & \textbf{2.75} & \textbf{92.71} & \textbf{96.38} & \textbf{82.12} & \textbf{91.66} & 13.83 & 58.65 \\
PDM ($\lambda=3$) & 4.36 & 2.95 & \underline{92.08} & 95.77 & 81.25 & 90.39 & 13.33 & 61.79 \\
PDM ($\lambda=5$) & 4.40 & 3.00 & 92.27 & 95.96 & 80.89 & 90.75 & \underline{12.97} & \textbf{56.85} \\
PDM ($\lambda=10$) & 4.49 & 3.07 & 91.60 & 95.35 & 80.21 & 89.93 & \textbf{12.96} & \underline{56.91} \\

\bottomrule
\end{tabular*}

\end{table}

Table~\ref{tab:lambda} studies the direction-matching weight $\lambda$ in PDM.
When $\lambda=0$, PDM reduces to positive-only matching. We find that
$\lambda=1$ provides the best overall control fidelity, including the lowest
keyframe MPJPE and the highest PCK@0.1. Larger values improve some
distributional-quality metrics but weaken control accuracy. We therefore use
$\lambda=1$ as the default.

\subsubsection{Effect of the Supervision Horizon}
\label{sec:opd_horizon}

\begin{table}[t]
\centering
\small
\setlength{\tabcolsep}{4pt}

\caption{\textbf{Effect of the OPD supervision horizon \(K\) in the pose-only ablation setting.} Wall-clock training time is reported in seconds per optimization step. \textbf{Bold} and \underline{underlined}
numbers denote the best and second-best results, respectively.
}
\label{tab:opd_horizon_ablation}

\begin{tabular*}{\textwidth}{@{\extracolsep{\fill}}l
cc
cc
cc
cc
c}
\toprule

& \multicolumn{2}{c}{MPJPE $\downarrow$}
& \multicolumn{2}{c}{PCK@0.2 $\uparrow$}
& \multicolumn{2}{c}{PCK@0.1 $\uparrow$}
& \multicolumn{2}{c}{Quality $\downarrow$}
& \multicolumn{1}{c}{Time $\downarrow$} \\
\cmidrule(lr){2-3}\cmidrule(lr){4-5}\cmidrule(lr){6-7}\cmidrule(lr){8-9}\cmidrule(lr){10-10}

Method
& All & Key
& All & Key
& All & Key
& FID & FVD
& s/step \\
\midrule

PDM ($K=1$) & 4.33 & 2.98 & 92.53 & 95.68 & 81.19 & 90.57 & \underline{13.70} & \textbf{54.41} & $\sim$\textbf{23} \\
PDM ($K=2$) & 4.45 & 3.11 & 92.18 & 95.57 & 80.78 & 90.51 & 13.89 & 60.63 & $\sim$\underline{38} \\
PDM ($K=4$) & \textbf{4.27} & \underline{2.86} & \underline{92.59} & \underline{95.98} & \underline{81.71} & \underline{91.13} & \textbf{13.58} & \underline{57.08} & $\sim$69 \\
PDM ($K=8$) & \textbf{4.27} & \textbf{2.75} & \textbf{92.71} & \textbf{96.38} & \textbf{82.12} & \textbf{91.66} & 13.83 & 58.65 & $\sim$132 \\
PDM ($K=50$) & 4.65 & 3.14 & 91.34 & 95.25 & 80.81 & 90.09 & 13.91 & 61.95 & $\sim$790 \\
\bottomrule
\end{tabular*}

\end{table}

Table~\ref{tab:opd_horizon_ablation} studies the number of supervised states
$K$ in the dense-to-sparse video application. Increasing $K$ raises training
cost from 23 seconds per step at $K=1$ to 790 seconds under full-trajectory
supervision ($K=50$). Among the evaluated settings, $K=8$ provides the
strongest control fidelity while remaining substantially more efficient than
full-trajectory supervision. We therefore use $K=8$ for the main video
experiments.

\section{Related Work}
\label{sec:related_work}

\subsection{Diffusion Model Distillation}
\label{sec:rw_diffusion_distillation}

Although our goal is not few-step generation, prior diffusion
distillation provides the closest context for understanding how teacher
supervision and classifier-free guidance are typically transferred to a
student. Most existing methods focus on reducing the number of denoising steps
required at inference. Progressive Distillation recursively compresses multiple
teacher steps into a single student step~\citep{salimans2022progressive}.
Consistency-based methods, including Consistency
Models~\citep{song2023consistency,kim2024consistency} and Latent Consistency
Models~\citep{luo2023latent}, train a model to produce consistent outputs across
states along the same probability-flow trajectory, enabling one- or few-step
generation. Distribution Matching Distillation (DMD) matches the student
distribution to the teacher through their score
functions~\citep{dmd}, while other approaches further combine diffusion
supervision with adversarial training~\citep{sauer2024adversarial,dmd2}.

Many of these methods also distill classifier-free guidance into the student.
Rather than retaining separate positive and negative branches at inference,
they use the CFG-composed teacher prediction as the distillation target, thereby
baking the effect of guidance into a single student
prediction~\citep{luo2023latent,dmd,collectionlora,liveavatar}. This design
improves sampling efficiency by avoiding two model evaluations for CFG and,
more broadly, transfers the strongly guided teacher distribution into a compact
few-step generator.

Our setting differs in both objective and purpose. We do not primarily seek to
compress a multi-step guided teacher into a CFG-free few-step student. Instead,
we study diffusion OPD in which the teacher and student each construct their own
guided prediction from positive and negative branches. Matching only their
final guided predictions is under-identified at the branch level. This
ambiguity becomes harmful when the teacher's negative branch contains
privileged information unavailable to the student, preventing the two branch
errors from being reduced jointly.

\subsection{On-Policy Distillation for Diffusion Models}
\label{sec:rw_diffusion_opd}

Recent work extends on-policy distillation to diffusion and flow models by
querying the teacher on states sampled from the current student. D-OPSD performs
on-policy self-distillation using privileged context, allowing a student to
learn from a stronger teacher evaluated along the student's own denoising
trajectory~\citep{dopsd}. Flow-OPD distills task-specialized flow-matching
teachers into a unified student through on-policy sampling and dense per-step
supervision~\citep{flowopd}. DiffusionOPD formulates this process through
transition-level KL divergence and shows that it reduces to prediction matching
under standard diffusion transitions~\citep{diffusionopd}. Concurrently,
DanceOPD treats generative capabilities as velocity fields and internalizes
operator-defined fields such as CFG into a single student prediction through
on-policy field matching~\citep{danceopd}. Unlike the branch-retaining setting
studied here, such a student no longer exposes separate positive and negative
predictions at inference.

Despite differences in derivation and optimization, these methods share the
same central principle: the student generates the trajectory, while the teacher
provides dense local supervision at student-visited states. Their studied
configurations, however, either collapse CFG into a single student prediction
or retain separate branches while sharing the same negative conditioning
between teacher and student. In the latter case, the negative-branch error need
not be exactly zero, but shared conditioning allows the two branch errors to be
reduced jointly. Existing work therefore does not examine how privileged
information in the teacher's negative branch changes the optimization of a
branch-retaining student.

Our work studies this missing regime. We show that CFG-composed matching permits
cross-branch error compensation and identify when this ambiguity becomes
harmful: under shared negative conditioning, both branch errors can decrease
jointly, whereas privileged negative conditioning can induce antagonistic
branch-error dynamics. We term this failure mode NBA. We further introduce branch-aware OPD objectives, with
PDM as the primary formulation, to prevent such error
compensation and improve robustness across inference guidance scales.

\section{Conclusions and Limitations}
\label{sec:conclusion_limitation}

We show that CFG-composed OPD is under-identified at the branch level, but that this ambiguity does not always cause failure. Under shared negative conditioning, the positive- and negative-branch errors can decrease jointly. When the teacher's negative branch contains privileged information unavailable to the student, however, joint error reduction can break down and naive matching induces antagonistic branch-error dynamics. We identify this failure mode as Negative Branch Asymmetry (NBA). To address it, we introduce Positive--Direction Matching (PDM), which separately supervises the positive prediction and CFG conditional direction. Contrasting image-domain studies validate when NBA emerges, while dense-to-sparse video control demonstrates that branch-aware supervision improves knowledge transfer and robustness across inference guidance scales.

Despite these results, the relative behavior of branch-aware objectives remains incompletely understood. PDM and IBM share the same branch-level zero-loss solution under non-degenerate weights, so PDM's observed advantage is currently empirical rather than theoretical. Future work may characterize their different optimization behavior and extend NBA analysis to other guidance mechanisms and teacher--student conditioning asymmetries.

\vspace{0.25cm}
\newpage
\bibliography{iclr2026_conference}
\bibliographystyle{iclr2026_conference}
\newpage
\section{Appendix}

\subsection{Additional Experimental Details}
\label{app:setup}

\subsubsection{Text Rendering Distillation}
\label{app:text_rendering_distillation}

\paragraph{Models.}
The base model is Stable Diffusion~3.5-Medium~\citep{sd3}. Both the teacher
and the student are LoRA adapters~\citep{lora} of rank $r=32$ and scale
$\alpha=64$, applied to the query/key/value and output projections of every
self- and cross-attention block in the MM-DiT transformer; all other weights are
frozen. We use the same OCR teacher model used in DiffusionOPD~\citep{diffusionopd}; the student is initialized from the base
model (identity adapter) and is the only module updated during distillation.

\paragraph{Objectives.}
We compare, under identical rollouts and hyper-parameters, the naive
CFG-composed objective, Positive-Direction Matching with $\lambda=2$ and the positive-only ablation. Only the pointwise objective differs across methods. The positive- and negative-branch errors reported in
Figure~\ref{fig:branch_dynamics} are per-transition $\ell_2$ norms
$\|\mathbf e_{\pm}\|_2$ in transition-mean space, averaged over the
supervised timesteps and logged every optimizer step.

\paragraph{Optimization.}
We use AdamW ($\beta_1{=}0.9$, $\beta_2{=}0.999$, $\epsilon{=}10^{-8}$, weight
decay $10^{-4}$) with a constant learning rate of $3\times10^{-4}$ and gradient
clipping at norm $1.0$, keeping an EMA of the trainable parameters (decay $0.9$,
updated every $8$ steps). Training runs on $8\times$H100 GPUs in fp16 mixed
precision with a per-GPU sampling batch of $3$ prompts, one image per prompt, and
$3$ sampling batches accumulated per update, giving an effective batch of
$8\times3\times3=72$ trajectories ($\times\,K$ transitions) per step. Each round
performs a single optimizer update; we train for $1000$ updates with seed $42$.

\paragraph{Data and reward.}
Prompts are drawn from the OCR text-rendering set ($19{,}652$ train /
$1018$ test) used in DiffusionOPD~\citep{diffusionopd}, where the target string $s$ to be rendered is given in quotes.
For evaluation, we read the text $\hat{s}$ from the generated image with PaddleOCR~\citep{cui2025paddleocr} and score
\begin{equation}
  R_{\text{OCR}} \;=\; 1 - \frac{\mathrm{Lev}(\hat{s}, s)}{\max(|\hat{s}|,\,|s|)}
  \;\in\; [0,1],
\end{equation}
where $\mathrm{Lev}(\hat{s},s)$ is the Levenshtein (edit) distance between the two
strings---the minimum number of single-character insertions, deletions, or
substitutions to turn $\hat{s}$ into $s$---and $|\cdot|$ denotes string length.
A score of $1$ means the rendered text is read back exactly; $0$ means no
characters match.

\paragraph{Evaluation.}
Unless noted, models are evaluated on the full $1018$-prompt test set with
$40$ denoising steps in fp16. For the guidance-scale generalization study we
sweep the inference guidance $\gamma\in\{1,2,3,4,4.5\}$ while keeping the
$\gamma_{\text{train}}=4.5$ training configuration fixed.

\subsubsection{Reference-Conditioned Distillation}
\label{app:ref_conditioned_distillation}
\paragraph{Data and task construction.}
Following the style-learning experiment in \citeauthor{dopsd}, we
construct a style-distillation setting spanning six distinct visual
styles, denoted by the placeholders [A]--[F]. For each style we provide four
training demonstrations~\citep{collectionlora,searchgen}, each pairing a text prompt with a reference image that
exposes the target style to the teacher, yielding $6 \times 4 = 24$ training
demonstrations in total. The demonstrations are organized by style: every
demonstration for style [A] shares the same target appearance, and likewise for
[B]--[F]. We train the LoRA for $2000$ gradient steps, drawing a batch of $16$
sampled records per gradient step; different visits use independently sampled
initial noise and denoising states, so repeated exposure to a demonstration
still yields stochastic training observations. For qualitative evaluation, we
construct $12$ held-out test prompts that invoke the same style placeholders
while changing the surrounding content, composition, or interactions. The goal
is to test whether the LoRA can distill the style knowledge exposed by the seen
demonstrations into its weights and, at inference, transfer a learned style to
unseen text prompts. The resulting evaluation tests whether the text-only
student internalizes the reference-supplied style through OPD rather than reproducing any
particular training scene.

\paragraph{Model and conditioning.}
We use the multi-step FLUX.2-klein-4B-base model~\citep{flux-2-2025} at
$512\!\times\!512$ resolution. Each reference image is converted to RGB,
bicubically resized to cover the target canvas, center-cropped, and normalized
to $[-1,1]$ before VAE encoding. The student follows the original text-to-image
interface and receives only the prompt. The teacher uses the same diffusion
backbone in its image-editing interface: it receives the student-visited noisy
latent together with the encoded reference-image latents, and its text condition
appends the instruction ``Make the overall style consistent with the reference image.'' For the
teacher's negative prediction, we replace the text with the null prompt while
retaining the reference-image latents. The student's negative prediction uses
the same null prompt without reference latents.

\paragraph{Parameterization and teacher update.}
We freeze the VAE, text encoder, and base diffusion-transformer weights and
optimize LoRA adapters with rank 64 and scaling parameter $\alpha=128$. The
adapters are applied to the query, key, value, and output projections of the
self- and context-attention modules, together with the corresponding
feed-forward projections in each transformer block. We maintain separate
student and teacher adapters initialized from the same weights. After each
student update, we update the teacher adapter with exponential moving average
decay $0.9999$. Teacher predictions are evaluated without gradients.

\paragraph{On-policy objectives.}
For each batch, the student initializes a trajectory from Gaussian noise under
the text-only condition. We use the native FLUX.2-klein~\citep{flux-2-2025} scheduler with a
30-step denoising grid. At every supervised state, the teacher and student construct their own
positive and negative predictions. The student rollout uses its CFG-composed
velocity at $\gamma_{\mathrm{train}}=2$, while the teacher is evaluated at the
same student-visited state.
We implement naive OPD by matching the teacher and student after CFG
composition. PDM instead matches the teacher's positive prediction and the CFG
conditional direction separately, with $\lambda=1$. All compared objectives use
the same base model, reference--prompt pairs, student rollout states, and
optimization budget.

\paragraph{Optimization.}
We train with AdamW using learning rate $10^{-4}$,
$\beta_1=0.9$, $\beta_2=0.999$, zero weight decay, and gradient-norm clipping at
1.0. Training uses eight processes with a per-process batch size of 2 and no
gradient accumulation, giving a global batch size of 16. We use bfloat16 mixed
precision for the trainable model and VAE, gradient checkpointing, and
DeepSpeed ZeRO-2. 

\paragraph{Inference and qualitative evaluation.}
At inference, we discard the teacher adapter and reference image; the student
receives only the held-out text prompt. We generate $512\!\times\!512$ images
with 40 denoising steps and evaluate guidance scales
$\gamma\in\{1,1.5,2,2.5\}$. For a given evaluation prompt, all objectives and
guidance scales use the same initial noise, isolating changes caused by the
training objective and CFG composition. We assess the resulting images
qualitatively by inspecting whether they preserve the defining visual knowledge
while following the changed context of the evaluation prompt and maintaining
generation quality. This experiment is an illustrative qualitative case study;
we do not interpret it as an aggregate estimate of performance.
\subsubsection{Dense-to-Sparse Video Benchmark}
\label{app:benchmark}

Our dense-to-sparse video benchmark is derived from OpenHumanVid
(OHV)~\citep{li2025openhumanvid}. Starting from the 12-part OHV release
($\approx91$K clips with DWPose annotations), we first filter clips
according to minimum length and motion magnitude. We then split videos
into non-overlapping 81-frame clips at $832\times480$ resolution while
preserving pixel alignment between RGB and control streams.

\paragraph{Difficulty stratification.}
To reduce evaluation bias toward easier samples~\citep{li2026overlaybench}, we compute a motion score based
on frame-to-frame differences of rendered DWPose skeletons. The clips
are divided into six difficulty buckets according to this score, and we
randomly sample 100 clips from each bucket to form the 600-clip test
set. The remaining clips are used for training and validation. Please refer to Table~\ref{tab:buckets} for more details.
\begin{table}[t]
\centering
\small
\begin{tabular}{@{}lccl@{}}
\toprule
Bucket & Motion score & \#Test & \#Train \\
\midrule
\texttt{diff1.5-2.0} & $[1.5, 2.0)$ & $100$ & $2{,}194$ \\
\texttt{diff2.0-2.5} & $[2.0, 2.5)$ & $100$ & $6{,}355$ \\
\texttt{diff2.5-3.5} & $[2.5, 3.5)$ & $100$ & $8{,}992$ \\
\texttt{diff3.5-5.0} & $[3.5, 5.0)$ & $100$ & $4{,}851$ \\
\texttt{diff5.0-7.0} & $[5.0, 7.0)$ & $100$ & $1{,}319$ \\
\texttt{diff7.0+}     & $[7.0, \infty)$ & $100$ & $364$ \\
\midrule
Total & & $600$ & $\approx\!24$K \\
\bottomrule
\end{tabular}
\caption{Difficulty-stratified benchmark. The test set draws $100$ clips from each
of six motion-difficulty buckets ($600$ total); motion score is the mean
frame-to-frame pixel difference of the rendered DWPose skeleton.}
\label{tab:buckets}
\end{table}

\subsubsection{Sparse Keyframe Control}
\label{app:keyframe}

All video experiments target the dense-to-sparse setting: the teacher receives the
control signal at \emph{every} frame, while the student receives it only at a
sparse set of pixel-space keyframes. Following the masked video-to-video (MV2V)
interface of the VACE~\citep{vace} base model, the student input frame stack $F$ and mask $M$
(over $81$ frames at $480\!\times\!832$, in the $[-1,1]$ range) are built as
\begin{itemize}
[leftmargin=*,topsep=0pt,itemsep=0pt,parsep=0pt,partopsep=0pt]
  \item $F[0]$ = the reference image (appearance anchor), with $M[0]=0$
        (given, not generated);
  \item $F[k]$ = the control frame at each pose keyframe $k\!\in\!\{20,40,60\}$;
  \item all remaining frames are blank ($-1$), with $M[\,\cdot\,]=1$
        (to be synthesized).
\end{itemize}
We use the uniform layout \texttt{pixel\_kf}$=\{0,20,40,60\}$ throughout: index $0$
is reserved for the reference image, so the model is conditioned on pose at three
interior keyframes out of $81$ frames. Keyframes are indexed in \emph{pixel} space;
under the VACE VAE ($4\times$ temporal, $8\times$ spatial compression) this maps
non-linearly to latent frames, so all masks are constructed in pixel space and
then compressed. This same sparse construction is used both inside the on-policy
rollout during training and at inference; the teacher instead receives the dense
per-frame control.

\subsubsection{Evaluation Metrics}
\label{app:metrics}

\begin{table}[t]
\centering
\small
\begin{tabular}{@{}ll@{}}
\toprule
\multicolumn{2}{l}{\emph{Base model \& adapter}} \\
\midrule
Base model            & Wan2.1-VACE-1.3B (frozen) \\
Adapter               & LoRA, rank $32$, $\alpha=32$ \\
LoRA target           & self/cross-attn $\{q,k,v,o\}$, ffn $\{0,2\}$ per VACE block \\
Precision             & bf16 autocast, gradient checkpointing \\
\midrule
\multicolumn{2}{l}{\emph{Optimization}} \\
\midrule
Optimizer             & AdamW, $\beta=(0.9,0.999)$, wd $=0$, $\epsilon=10^{-8}$ \\
Learning rate         & $1\times10^{-4}$ (constant) \\
Training steps        & $2000$ \\
Devices / batch       & $8$ GPUs (DDP), per-GPU batch $1$ \\
\midrule
\multicolumn{2}{l}{\emph{Diffusion \& rollout}} \\
\midrule
Solver / schedule     & UniPC, $50$ steps, \texttt{shift}$=16$ \\
OPD rollout $K$       & $8$ active steps (``$8$-of-$50$'') \\
Guidance scale $\gamma$    & $5.0$ \\
\midrule
\multicolumn{2}{l}{\emph{Data}} \\
\midrule
Resolution / length   & $480\!\times\!832$, $81$ frames \\
Keyframes             & pixel $\{0,20,40,60\}$ (0 = reference) \\
Train / val / test    & $\approx\!24$K / $24$ / $600$ clips \\
\bottomrule
\end{tabular}
\caption{Training and inference hyperparameters, shared across all distillation
objectives.}
\label{tab:hparams}
\end{table}

All methods are evaluated independently on the 600-clip test set.
Scores are computed per clip and averaged over the dataset.

\paragraph{Control fidelity (per modality).}
For each control modality we \emph{re-extract} the control signal from the
generated video with the corresponding annotator and compare it to the control
that was fed in.
\begin{itemize}
[leftmargin=*,topsep=0pt,itemsep=0pt,parsep=0pt,partopsep=0pt]
  \item \textbf{Pose.} We run DWPose~\citep{dwpose} on the generated video and use the $18$
        body joints, in normalized $[0,1]$ image coordinates. A joint is counted
        only when both the GT and predicted confidence exceed $0.3$; frames with
        fewer than $3$ valid joints are skipped. We report
        \emph{MPJPE} (mean per-joint $\ell_2$ error in normalized coordinates) and
        \emph{PCK@$\tau$} for $\tau\!\in\!\{0.2,0.1\}$, where a joint is correct if
        its error is below $\tau\cdot s$ and $s$ is the per-frame bounding-box size
        $s=\max(\Delta x, \Delta y)$ of the GT skeleton. Metrics are reported over
        \emph{all} frames and over the \emph{keyframe} subset separately.
  \item \textbf{Depth.} We re-extract depth from the generated frames with
        DPT-Hybrid (MiDaS)~\citep{midas} and compare against the dense depth control (same
        annotator). Since these are \emph{relative} depths, we remove the scale/shift
        ambiguity with a closed-form least-squares affine alignment
        $a\cdot\hat d + b \approx d$ and report the affine-invariant
        \emph{SI-RMSE} (primary), the raw RMSE, and the Pearson correlation.
  \item \textbf{Scribble.} We re-extract neural line-art from the generated frames
        with the VACE scribble annotator and compare against the dense line-art
        control. Line pixels are binarized as deviations from the per-frame median
        ($>\!40$/$255$), and we report a tolerant boundary
        \emph{F1} (precision/recall) with a dilation tolerance of $2$ pixels.
\end{itemize}

\paragraph{Distributional quality.}
Independent of control adherence, we measure perceptual and temporal realism
against the GT distribution: \emph{FID} (clean-fid, InceptionV3 features, $8$ frames
per clip) captures per-frame appearance, and \emph{FVD} (I3D pretrained on
Kinetics-400, $16$ frames linspace-sampled per clip) captures spatio-temporal
consistency. 

\subsubsection{Implementation and Hyperparameters}
\label{app:hparams}

All methods fine-tune the same Wan2.1-VACE-1.3B~\citep{vace} base model with LoRA~\citep{lora}
adapters (the base weights are frozen), so differences are attributable to the
training objective alone. LoRA (rank $32$, $\alpha\!=\!32$) is applied to the
query/key/value/output projections of both self- and cross-attention and the two
feed-forward linears within every VACE block. We train with AdamW
($\text{lr}=10^{-4}$, $\beta=(0.9,0.999)$, no weight decay) for $2000$ steps on
$8$ GPUs (DDP, per-GPU batch $1$), under bf16 mixed precision with gradient
checkpointing. Inference and the on-policy rollout use the UniPC solver~\citep{zhao2023unipc} on a
$50$-step schedule (\texttt{shift}$=16$). All models are
distilled and deployed at classifier-free guidance scale $\gamma=5$. Videos are $81$
frames at $480\!\times\!832$. Table~\ref{tab:hparams} lists the full configuration.

\paragraph{Off-policy distillation baseline.}
The off-policy baseline uses the same teacher--student prediction-matching
loss as OPD but evaluates it on standard data-derived diffusion states. Given
a ground-truth video latent $\mathbf{x}_0$, we sample $t$ and
$\boldsymbol{\epsilon}\sim\mathcal{N}(\mathbf{0},\mathbf{I})$ and construct
$\mathbf{x}_t^{\mathrm{off}}
=\alpha_t\mathbf{x}_0+\sigma_t\boldsymbol{\epsilon}$.
While SFT matches the student prediction at this state to the ground-truth
velocity target, the off-policy baseline replaces that target with the teacher
prediction under dense control. The student is evaluated at the same state
under sparse control. It therefore differs from OPD only in the state
distribution, using noised ground-truth latents instead of student-visited
states.

\subsection{More Qualitative Results}
\label{app:more_qual_res}

\begin{figure}[t]
    \centering
    \includegraphics[width=1\linewidth]{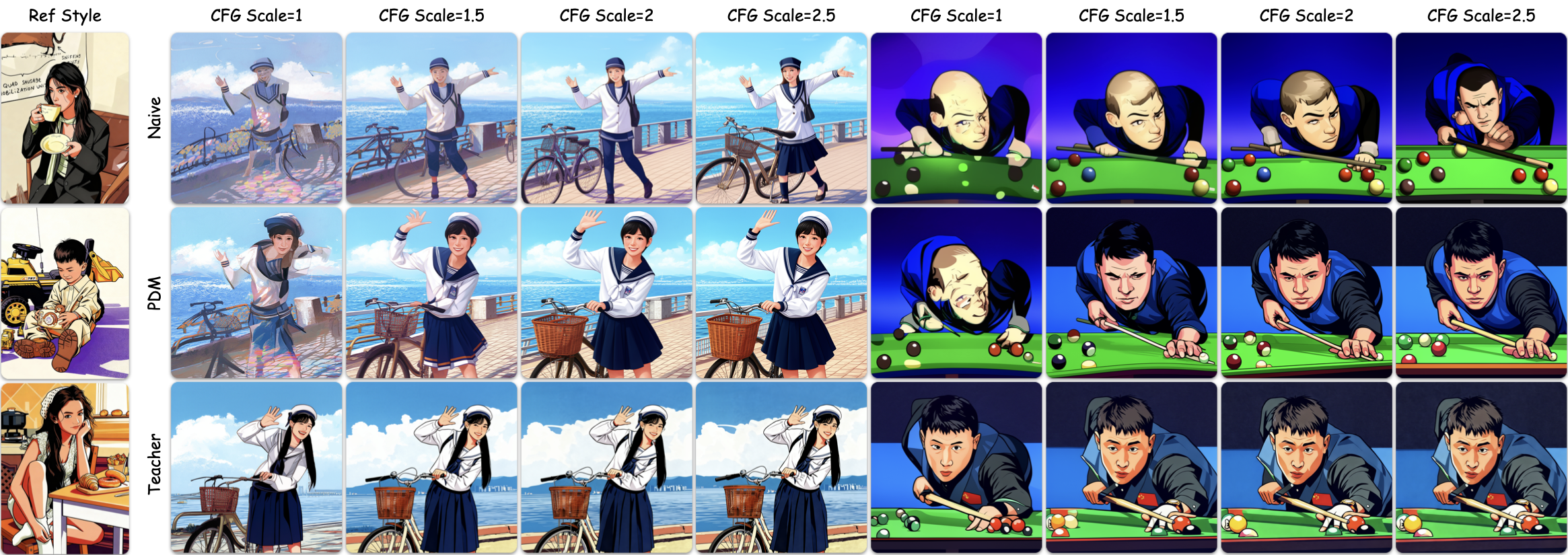}
\caption{\textbf{Additional qualitative results for reference-conditioned
distillation.} The left column shows reference-style exemplars, followed by
outputs for two held-out prompts across inference guidance scales. Text-only
students are trained at $\gamma_{\mathrm{train}}=2$ using naive matching or
PDM, and the reference-conditioned teacher is shown for comparison. PDM more
consistently follows the teacher and preserves the reference-supplied style,
whereas naive matching exhibits stronger appearance drift and visual artifacts,
particularly at $\gamma=1$.}
\label{fig:reference_cfg_appendix}
\end{figure}

\begin{figure}[t]
    \centering
    \includegraphics[width=1\linewidth]{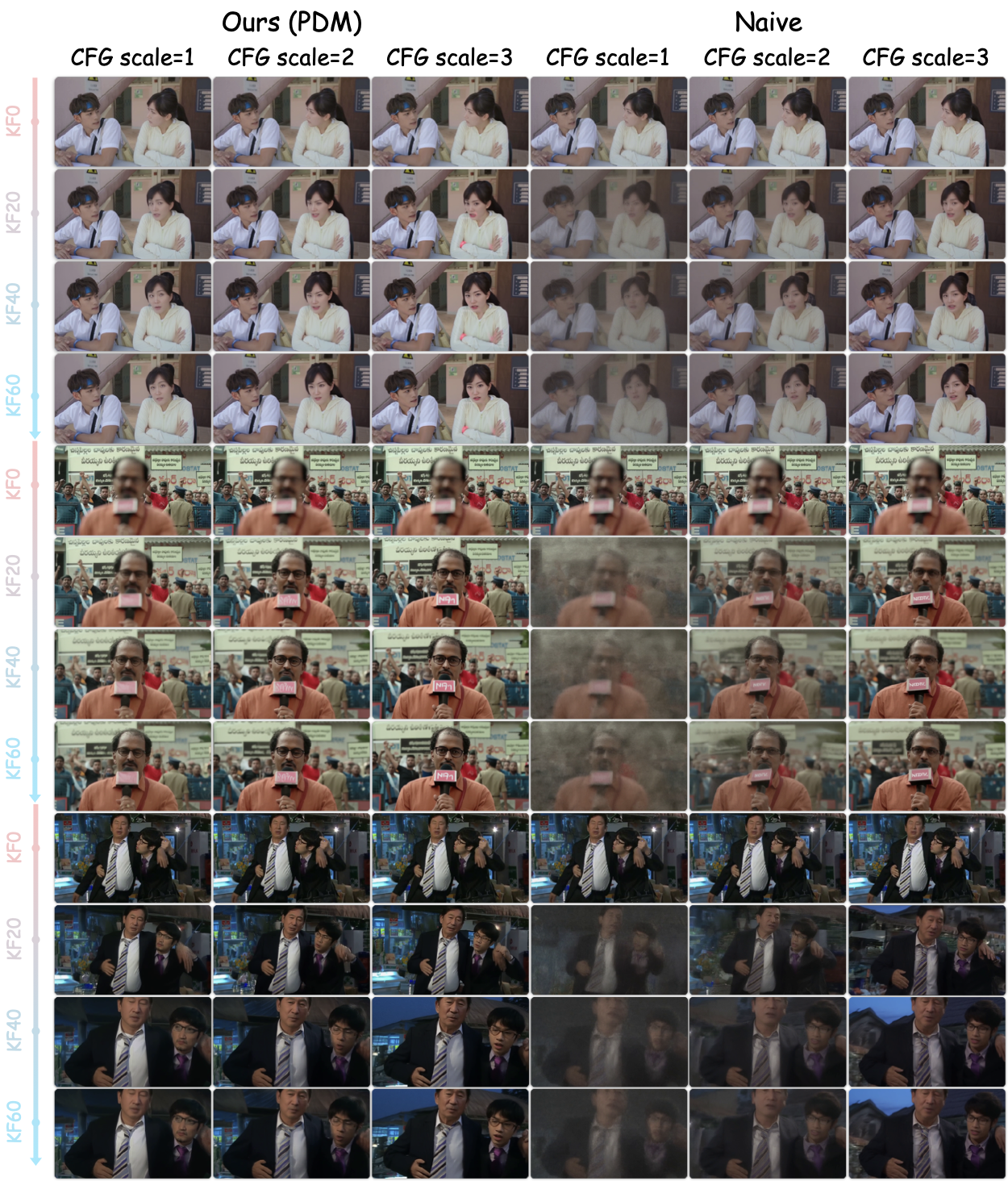}
    \caption{\textbf{Additional qualitative visualization of NBA in
dense-to-sparse video control.} Models are trained at
$\gamma_{\mathrm{train}}=5$ and evaluated at lower inference guidance scales.
Naive CFG-based OPD exhibits severe blurring and structural degradation as
$\gamma_{\mathrm{test}}$ moves away from the training configuration, revealing
its sensitivity to a specific CFG composition. PDM consistently preserves
subject appearance and video structure across guidance scales by separately
matching the positive prediction and CFG conditional direction.}
    \label{fig:d2s_nba_app}
\end{figure}

Figures~\ref{fig:reference_cfg_appendix} and~\ref{fig:d2s_nba_app}
provide additional qualitative evidence of distillation-induced CFG sensitivity
under privileged conditioning. In reference-conditioned distillation, PDM more
closely follows the teacher and preserves the reference-supplied style across
guidance scales, whereas naive matching exhibits appearance drift and visual
artifacts. In dense-to-sparse video control, PDM likewise maintains subject
appearance and controlled structure as the inference scale changes, while naive
matching degrades substantially. These results further demonstrate the excess
guidance sensitivity associated with NBA.

\begin{figure}[t]
    \centering
    \includegraphics[width=1\linewidth]{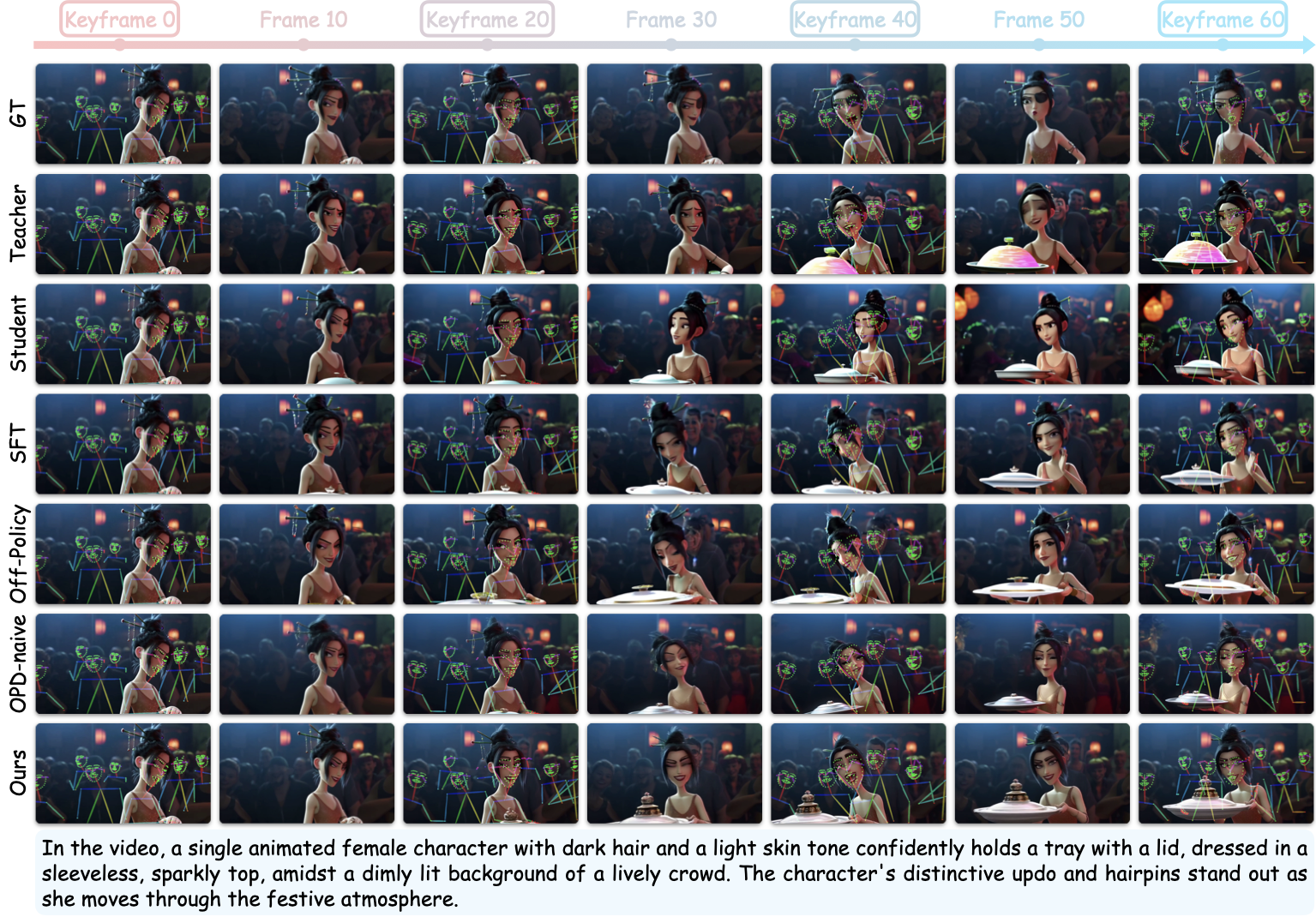}
    \caption{\textbf{Qualitative comparison on dense-to-sparse pose control distillation
under the default guidance scale $\gamma=5$.} The teacher receives dense
pose guidance, while the student is conditioned only on sparse
keyframes. PDM achieves more consistent
structure and visual quality compared with other distillation
objectives.}
    \label{fig:qual_res_pose_app}
\end{figure}

\begin{figure}[t]
    \centering
    \includegraphics[width=1\linewidth]{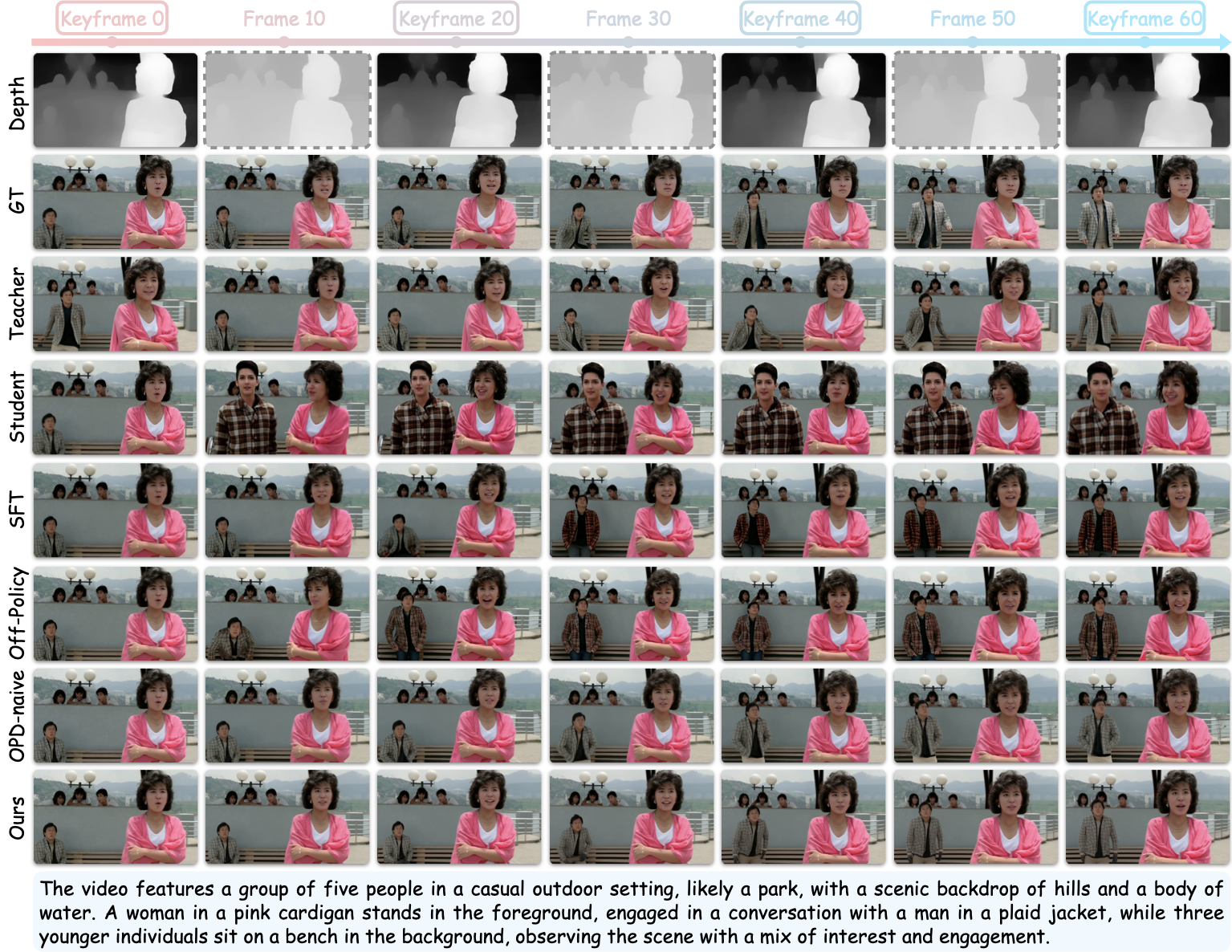}
    \caption{\textbf{Qualitative comparison on dense-to-sparse depth control distillation
under the default guidance scale $\gamma=5$.} The teacher receives dense
depth guidance, while the student is conditioned only on sparse
keyframes. PDM achieves more consistent
structure and visual quality compared with other distillation
objectives.}
    \label{fig:qual_res_depth_app}
\end{figure}

\begin{figure}[t]
    \centering
    \includegraphics[width=1\linewidth]{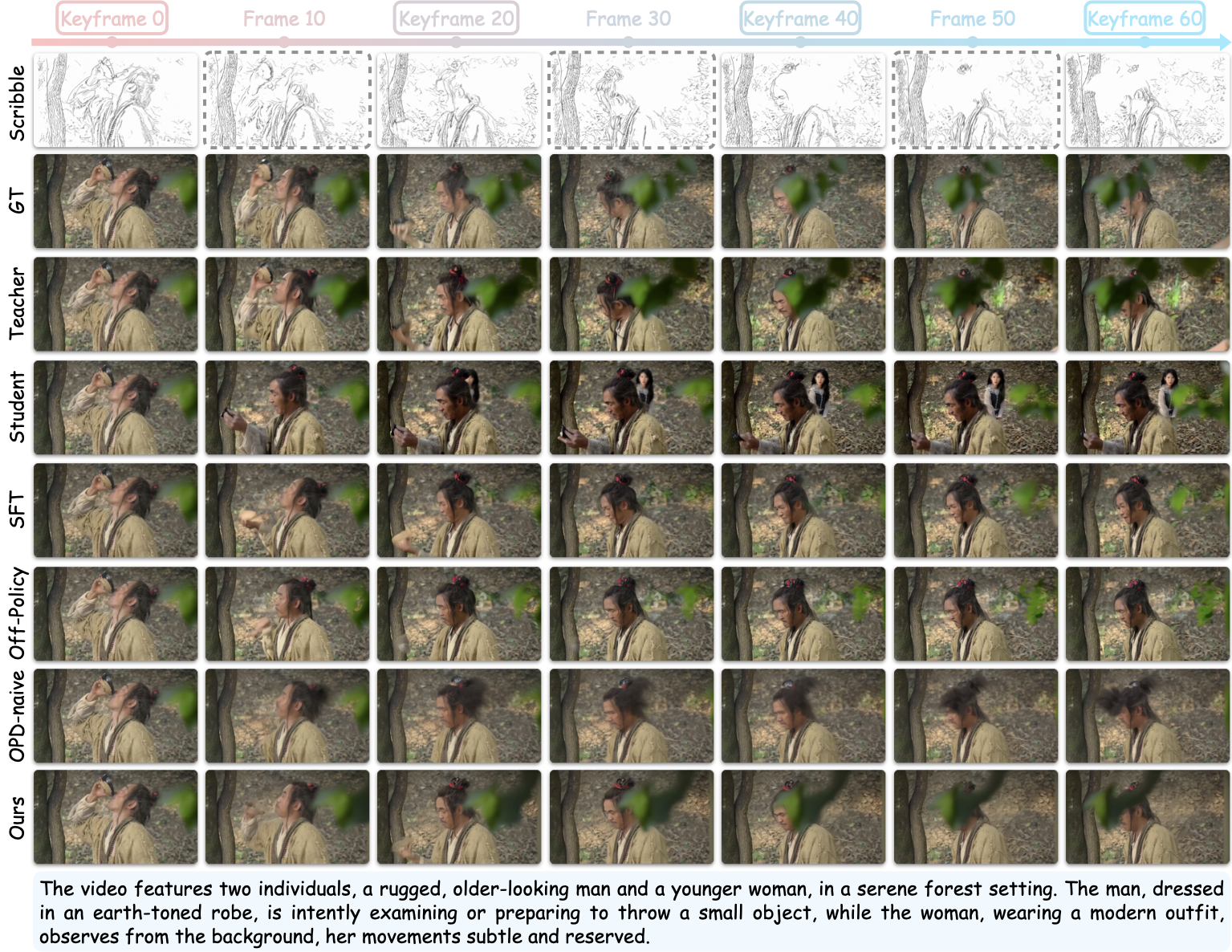}
    \caption{\textbf{Qualitative results on dense-to-sparse scribble control distillation with
$\gamma=5$.} Compared with other distillation objectives,
Positive--Direction Matching better preserves the scribble-guided
structure and generates more consistent visual results under sparse
control supervision.}
    \label{fig:qual_res_scribble_app}
\end{figure}

Figures~\ref{fig:qual_res_pose_app},~\ref{fig:qual_res_depth_app}
and~\ref{fig:qual_res_scribble_app} extend the qualitative comparison to pose, depth
and scribble control under the default guidance scale. Across all three modalities,
PDM follows the sparse structural conditions more faithfully while maintaining
coherent visual content. These results show that the benefit of branch-aware
supervision is consistent across different forms of conditioning asymmetry.

\end{document}